\documentclass[final]{cvpr}

\usepackage{times}
\usepackage{epsfig}
\usepackage{graphicx}
\usepackage{amsmath}
\usepackage{amssymb}
\usepackage{subfigure}
\usepackage{amsthm}
\usepackage{mathrsfs}
\usepackage{color}
\usepackage{multirow}
\usepackage{marvosym}

\usepackage{makecell}
\usepackage{textcomp}
\usepackage{array} 
\usepackage{longtable}
\usepackage{booktabs}
\usepackage{colortbl}
\definecolor{Gray}{gray}{.9}
\DeclareMathOperator*{\argmax}{argmax}

\usepackage[pagebackref=true,breaklinks=true,colorlinks,bookmarks=true]{hyperref}

\begin{document}

\title{Robust Reference-based Super-Resolution via $C^{2}$-Matching}

\author{Yuming Jiang$^{1}$
\quad
Kelvin C.K. Chan$^{1}$
\quad
Xintao Wang$^{2}$
\quad
Chen Change Loy$^{1}$
\quad
Ziwei Liu\textsuperscript{1\Letter}\\
$^{1}$S-Lab, Nanyang Technological University
\quad
$^{2}$Applied Research Center, Tencent PCG
\\
{\tt\small \{yuming002, chan0899, ccloy, ziwei.liu\}@ntu.edu.sg \hspace{5pt} xintao.wang@outlook.com}\\
}

\maketitle

\begin{abstract}
   Reference-based Super-Resolution (Ref-SR) has recently emerged as a promising paradigm to enhance a low-resolution (LR) input image by introducing an additional high-resolution (HR) reference image. 
   Existing Ref-SR methods mostly rely on \textbf{implicit correspondence matching} to borrow HR textures from reference images to compensate for the information loss in input images.
   However, performing local transfer is difficult because of two gaps between input and reference images: the transformation gap (\eg scale and rotation) and the resolution gap (\eg HR and LR).   
   To tackle these challenges, we propose $C^{2}$-Matching in this work, which produces \textbf{explicit robust matching crossing transformation and resolution}.
   1) For the transformation gap, we propose a contrastive correspondence network, which learns transformation-robust correspondences using augmented views of the input image.
   2) For the resolution gap, we adopt a teacher-student correlation distillation, which distills knowledge from the easier HR-HR matching to guide the more ambiguous LR-HR matching.   
   3) Finally, we design a dynamic aggregation module to address the potential misalignment issue.
   In addition, to faithfully evaluate the performance of Ref-SR under a realistic setting, we contribute the Webly-Referenced SR (WR-SR) dataset, mimicking the practical usage scenario.  
   Extensive experiments demonstrate that our proposed $C^{2}$-Matching significantly outperforms state of the arts by over 1dB on the standard CUFED5 benchmark.
   Notably, it also shows great generalizability on WR-SR dataset as well as robustness across large scale and rotation transformations~\footnote{Codes and datasets are available at \url{https://github.com/yumingj/C2-Matching}.}.
   
\end{abstract}

\vspace{-0.37cm}
\section{Introduction}

\begin{figure}[t]
  \begin{center}
      \includegraphics[width=1.0\linewidth]{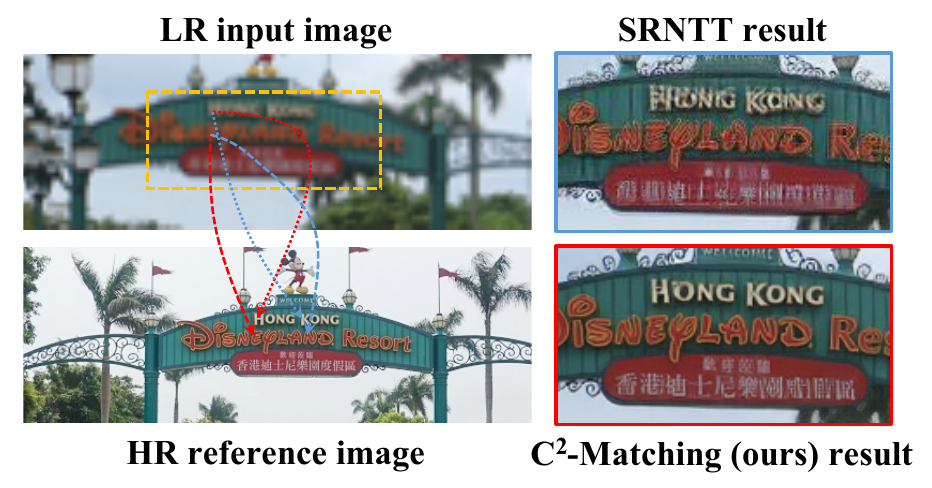}
  \end{center}
  \vspace{-13pt}
  \caption{\textbf{\textit{Cross transformation} and \textit{Cross resolution} matching are performed in our $C^{2}$-Matching}. Our proposed $C^{2}$-Matching successfully transfers the HR details of the reference image by finding more accurate correspondences. The correspondences found by our method are marked in red and the correspondences found by SRNTT \cite{zhang2019image} are marked in blue.}
  \vspace{-0.37cm}
  \label{teaser}
\end{figure}

Reference-based Super-Resolution (Ref-SR) \cite{zheng2018crossnet, zhang2019image, yang2020learning, Shim_2020_CVPR} has attracted substantial attention in recent years. Compared to Single-Image-Super-Resolution (SISR) \cite{dong2015image, kim2016accurate, kim2016deeply, lim2017enhanced, shi2016real, dai2019second}, where the only input is a single low-resolution (LR) image, Ref-SR super-resolves the LR image with the guidance of an additional high-resolution (HR) reference image. Textures of the HR reference image are transferred to provide more fine details for the LR image.

The key step in texture transfer for Ref-SR is to find correspondences between the input image and the reference image. Existing methods \cite{zhang2019image, yang2020learning, xiefeature} perform correspondence matching implicitly. Their correspondences are computed based on the content and appearance similarities, which are then embedded into the main framework.
However, it is a difficult task to accurately compute the correspondences under real-world variations due to two major challenges: \textbf{1)} the underlying transformation gap between input images and reference images; \textbf{2)} the resolution gap between input images and reference images. 
In Ref-SR, same objects or similar texture patterns are often present in both input images and reference images, but their appearances vary due to scale and rotation transformations. In this case, correspondences computed purely by appearance are inaccurate, leading to an unsatisfactory texture transfer.
For the resolution gap, due to the imbalance in the amount of information contained in an LR input image and an HR reference image, the latter is often downsampled (to an LR image) to match the former (in resolution). The downsampling operation inevitably results in information loss, hampering the search for accurate correspondences, especially for the fine-texture regions.

To address the aforementioned challenges, we propose $C^{2}$-matching for Robust Reference-based Super-Resolution, where \textbf{\textit{Cross transformation}} and \textbf{\textit{Cross resolution}} matching are explicitly performed. 
To handle the transformation gap, a contrastive correspondence network is proposed to learn transformation-robust correspondences between input images and reference images. Specifically, we employ an additional triplet margin loss to minimize the distance of point-wise features before and after transformations while maximizing the distance of irrelevant features. Thus, the extracted feature descriptors are more robust to scale and rotation transformations, and can be used to compute more accurate correspondences. 

As for the resolution gap, inspired by knowledge distillation, we propose a teacher-student correlation distillation. 
We train the teacher contrastive correspondence network for HR-HR matching. Since the teacher network takes two HR images as input, it is better at matching the regions with complicated textures. Thus, the knowledge of the teacher model can be distilled to guide the more ambiguous LR-HR matching.
The teacher-student correlation distillation enables the contrastive correspondence network to compute correspondences more accurately for texture regions.

After obtaining correspondences, we then fuse the information of reference images through a dynamic aggregation module to transfer the HR textures.
With $C^{2}$-Matching, we achieve over 1dB improvement on the standard CUFED5 dataset.
As shown in Fig.~\ref{teaser}, compared to SRNTT \cite{zhang2019image}, our $C^{2}$-Matching finds more accurate correspondences (marked as red dotted lines) and thus has a superior restoration performance.

To facilitate the evaluation of Ref-SR tasks in a more realistic setting, 
we contribute a new dataset named Webly-Reference SR (WR-SR) dataset. In real-world applications, given an LR image, users may find its similar HR reference images through some web search engines. Motivated by this, for every input image in WR-SR, we search for its reference image through Google Image. The collected WR-SR can serve as a benchmark for real-world scenarios.

To summarize, our main contributions are:
\textbf{1)} To mitigate the transformation gap, we propose the contrastive correspondence network to compute correspondences more robust to scale and rotation transformations.
\textbf{2)} To bridge the resolution gap, a teacher-student correlation distillation is employed to further boost the performance of student LR-HR matching model with the guidance of HR-HR matching, especially for fine texture regions.
\textbf{3)} We contribute a new benchmark dataset named Webly-Referenced SR (WR-SR) to encourage a more practical application in real scenarios. 

\section{Related Work}

\noindent\textbf{Single Image Super-Resolution.} Single Image Super-Resolution (SISR) aims to recover the HR details of LR images. The only input to the SISR task is the LR image. Dong \etal \cite{dong2015image} introduced deep learning into SISR tasks by formulating the SISR task as an image-to-image translation problem. 
Later, SR networks had gone deeper with the help of residual blocks and attention mechanisms \cite{shi2016real, lim2017enhanced, kim2016deeply, zhang2018image, zhang2018residual, ledig2017photo, dai2019second, zhou2020cross}. However, the visual quality of the output SR images did not improve. The problem was the mean square error (MSE) loss function. In order to improve the perceptual quality, perceptual loss \cite{johnson2016perceptual, sajjadi2017enhancenet}, generative loss and adversarial loss \cite{ledig2017photo} were introduced into the SR network \cite{wang2018esrgan, zhang2019ranksrgan}. 
Knowledge distillation framework is also explored to improve the SR performance in \cite{gao2018image, lee2020learning}.

\noindent\textbf{Reference-based Image Super-Resolution.} Different from SISR, where no additional information is provided, the Reference-based Image Super-Resolution (Ref-SR) task \cite{yantowards, zhangtexture, zheng2018crossnet} super-resolves input images by transferring HR details of reference images. 
Patch-Match method \cite{barnes2009patchmatch} was employed in \cite{zhang2019image, yang2020learning} to align input images and reference images. SRNTT \cite{zhang2019image} performed correspondence matching based on the off-the-shelf VGG features \cite{simonyan2014very}. Recently, \cite{yang2020learning, xiefeature} used learnable feature extractors, which were trained end-to-end accompanied with the main SR network. Even with the learnable feature extractors, the correspondences were computed purely based on contents and appearances. 
Recent work \cite{Shim_2020_CVPR} introduced Deformable Convolution Network (DCN) \cite{dai2017deformable, zhu2019deformable} to align input images and reference images. Inspired by previous work \cite{wang2019edvr, chan2021understand}, we also propose a similar module to handle the potential misalignment issue.

\noindent\textbf{Image Matching.} Scale Invariant Feature Transform (SIFT) \cite{lowe1999object} extracted local features to perform matching. With the advance of convolution neural networks (CNN), feature descriptors extracted by CNN were utilized to compute correspondences \cite{dusmanu2019d2, wiles2020d2d, duggal2019deeppruner}. Recently, SuperPoint \cite{detone2018superpoint} was proposed to perform image matching in a self-supervised manner, and graph neural network was introduced to learn feature matching \cite{sarlin2020superglue}. Needle-Match \cite{lotan2016needle} performed image matching in a more challenging setting, where two images used for matching are degraded. Different from the aforementioned methods dealing with two images of the same degradation, in our task, we focus on cross resolution matching, \ie matching between one LR image and one HR image. 
\begin{figure*}
   \begin{center}
      \includegraphics[width=1.0\linewidth]{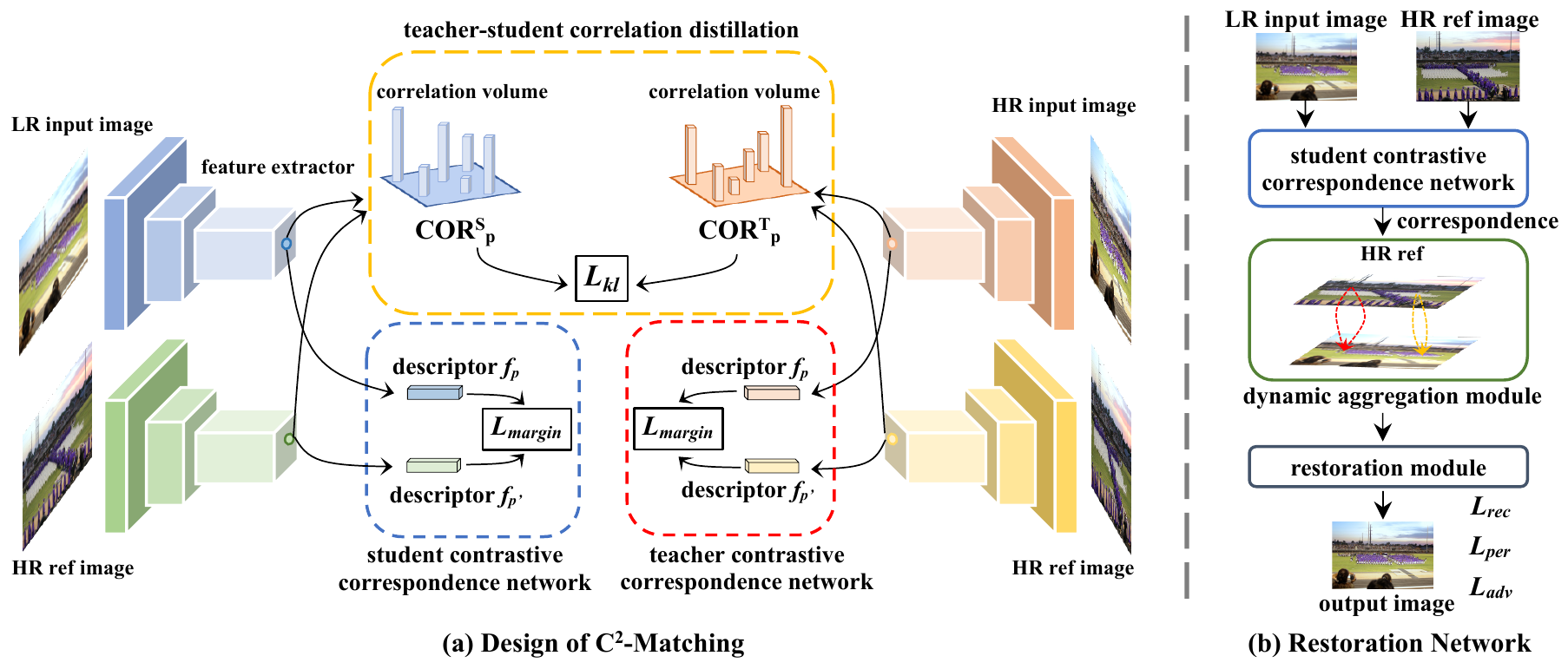}
   \end{center}
  \vspace{-10pt}
   \caption{\textbf{(a) The overview of our proposed $C^{2}$-Matching.} The contrastive correspondence network is designed for transformation-robust correspondence matching. The student contrastive correspondence network takes both the LR input image and HR reference image (the transformed version of the HR input image serves as the HR reference image during training) as input. The descriptors before and after transformations are pushed closer while distances of the irrelevant descriptors are maximized by $L_{margin}$. To enable the student LR-HR contrastive correspondence network to perform correspondence matching better on highly textured regions, we embed a teacher-student correlation distillation process to distill the knowledge of the easier HR-HR teacher matching network to the student model by $L_{kl}$. \textbf{(b) The overall pipeline of restoration network.}  The correspondences are first computed by the trained student contrastive correspondence network, after which the correspondences are used for subsequent dynamic aggregation module and restoration module.}
   \label{framework_illustration}
\end{figure*}

\section{Our Approach}

The overview of our proposed $C^{2}$-Matching is shown in Fig.~\ref{framework_illustration}(a). The proposed $C^{2}$-Matching consists of two major parts: 1) Contrastive Correspondence Network and 2) Teacher-Student Correlation Distillation. The contrastive correspondence network learns transformation-robust correspondence matching; the teacher-student correlation distillation transfers HR-HR matching knowledge to LR-HR matching for a more accurate correspondence matching on texture regions. The correspondences obtained from the contrastive correspondence network are used for the subsequent restoration network (Fig.~\ref{framework_illustration}(b)). In the restoration network, to better aggregate the information of reference images, a dynamic aggregation module is proposed to handle the underlying misalignment problem. Then the aggregated features are used for the later restoration task.

\subsection{Contrastive Correspondence Network}

To transfer textures from reference images, correspondences should first be computed between input images and reference images, \ie specify the location of similar regions in reference images for every region in input images.

Existing methods \cite{zhang2019image, yang2020learning, xiefeature} computed correspondences according to the content and appearance similarities between degraded input images and reference images. For example, Zhang \etal \cite{zhang2019image} used VGG features \cite{simonyan2014very} for the correspondence matching while \cite{yang2020learning, xiefeature} used features trained end-to-end together with the downstream tasks. 
The drawback of this scheme is that it cannot handle the matching well if there are scale and rotation transformations between input images and reference images. An inaccurate correspondence would lead to an imperfect texture transfer for restoration.
In this paper, we propose a learnable contrastive correspondence network to extract features that are robust to scale and rotation transformations. 

In contrastive correspondence network, we deal with the correspondence matching between LR input images and HR reference images. Since the resolutions of the input image and the reference image are different, we adopt two networks with the same architecture but non-shared weights for feature extractions. The architecture of feature extractors will be explained in supplementary files.

For training, we synthesize HR reference images by applying homography transformations to original HR input images. By doing so, for every position $p$ in the LR input image $I$, we can compute its ground-truth correspondence point $p^\prime$ in the transformed image $I^\prime$ according to the homography transformation matrix. 
We regard point $p$ and its corresponding point $p^\prime$ as a positive pair. During optimization, we push the distances between feature representations of positive pairs closer, while maximizing the distances between other irrelevant but confusing negative pairs (defined as Eq.~\eqref{negative_pair_def}). Similar to \cite{dusmanu2019d2}, we use the triplet margin ranking loss as follows:
\begin{equation}
   L_{margin} = \frac{1}{N}\sum_{p\in {I}}\max(0, {m + \mathrm{Pos}(p) - \mathrm{Neg}(p)}),
\label{loss_margin}
\end{equation}
where $N$ is the total number of points in image $I$ and $m$ is the margin value.

The positive distance $\mathrm{Pos}(p)$ between the descriptor $f_{p}$ of position $p$ and its corresponding descriptor $f_{p^\prime}$ is defined as follows:
\begin{equation}
   \mathrm{Pos}(p) = \left \| f_{p} - f_{p^\prime} \right \|_{2}^{2}.
\end{equation}

As for the calculation of negative distance $\mathrm{Neg}(p)$, to avoid easy negative samples dominating the loss, we only select the hardest sample. The negative distance is defined as follows:
\begin{equation}
   \begin{split}
    \mathrm{Neg}(p) = \min( \min_{k \in I^\prime, \left \| k - p^\prime \right \|_\infty  > T  } \left \| f_{p} - f_{k} \right \|_{2}^{2},\\
   \min_{k \in I,  \left \| k - p \right \|_\infty > T } \left \| f_{p^\prime} - f_{k} \right \|_{2}^{2}),
   \end{split}
\label{negative_pair_def}
\end{equation}
where $T$ is a threshold to filter out neighboring points of the ground-truth correspondence point.

Thanks to the triplet margin ranking loss and the transformed versions of input image pairs, the contrastive correspondence network can compute correspondences more robust to scale and rotation transformations. 
Once trained, the original LR input image and the HR reference image are fed into the contrastive correspondence network to compute correspondences. 

\subsection{Teacher-Student Correlation Distillation}

Since a lot of information is lost in LR input images, correspondence matching is difficult, especially for highly textured regions. 
Matching between two HR images has a better performance than LR-HR matching.
To mitigate the gap, we employ the idea of knowledge distillation \cite{hinton2015distilling} into our framework. Traditional knowledge distillation tasks \cite{hinton2015distilling, liu2019structured} deal with model compression issues, while we aim to transfer the matching ability of HR-HR matching to LR-HR matching. Thus, different from the traditional knowledge distillation models that have the same inputs but with different model capacities, in our tasks, the teacher HR-HR matching model and the student LR-HR matching have the exact same designs of architecture, but with different inputs.

The distances between the descriptors of HR input images and reference images provide additional supervision for knowledge distillation. Thus, we propose to push closer the correlation volume (a matrix represents the distances between descriptors of input images and reference images) of teacher model to that of student model. 
For an input image, we have $N$ descriptors, and its reference image has $M$ descriptors.
By computing correlations between descriptors of input images and reference images, we can obtain an $N \times M$ matrix to represent the correlation volume, and view it as a probability distribution by applying a softmax function with temperature $\tau$ over it. To summarize, the correlation of the descriptor of input image at position $p$ and the descriptor of reference image at position $q$ is computed as follows:
\begin{equation}
  \mathrm{cor}_{pq} = \frac{e^{\frac{f_p}{\left \| f_p \right \|} \cdot \frac{f_q}{\left \| f_q \right \|} /\tau }}{\sum_{k \in I^\prime} e^{\frac{f_p}{\left \| f_p \right \|} \cdot \frac{f_k}{\left \| f_k \right \|} /\tau }}.
  \label{correlation_equa}
\end{equation}

By computing the correlations $\mathrm{cor}_{pq}$ for every pair of descriptor $p$ and $q$, 
we can obtain the correlation volume. We denote $\mathrm{COR}^{T}$ and $\mathrm{COR}^{S}$ as the teacher correlation volume and student correlation volume, respectively.
For every descriptor $p$ of input image, the divergence of teacher model's correlation and student model's correlation can be measured by Kullback Leibler divergence as follows:
\begin{equation}
\begin{split}
   \mathrm{Div}_{p} = \mathrm{KL}(\mathrm{COR}_{p}^{T} || \mathrm{COR}_{p}^{S}) \\  = \sum_{k \in I^\prime} \mathrm{cor}_{pk}^{T}\: log(\frac{\mathrm{cor}_{pk}^{T}}{\mathrm{cor}_{pk}^{S}}).
\end{split}
\end{equation}

The correlation volume contains the knowledge of relationship between descriptors.
By minimizing the divergence between two correlation volumes, the matching ability of teacher model can be transferred to the student model. This objective is defined as follows:
\begin{equation}
   L_{kl} = \frac{1}{N}\sum_{p \in I} \mathrm{Div}_{p}.
\end{equation}

With the teacher-student correlation distillation, the total loss used for training the contrastive correspondence network is:
\begin{equation}
   L = L_{margin} + \alpha _{kl} \cdot L_{kl},
\end{equation}
where $\alpha _{kl}$ is the weight for the KL-divergence loss.

\begin{figure*}
   \begin{center}
      \includegraphics[width=1.0\linewidth]{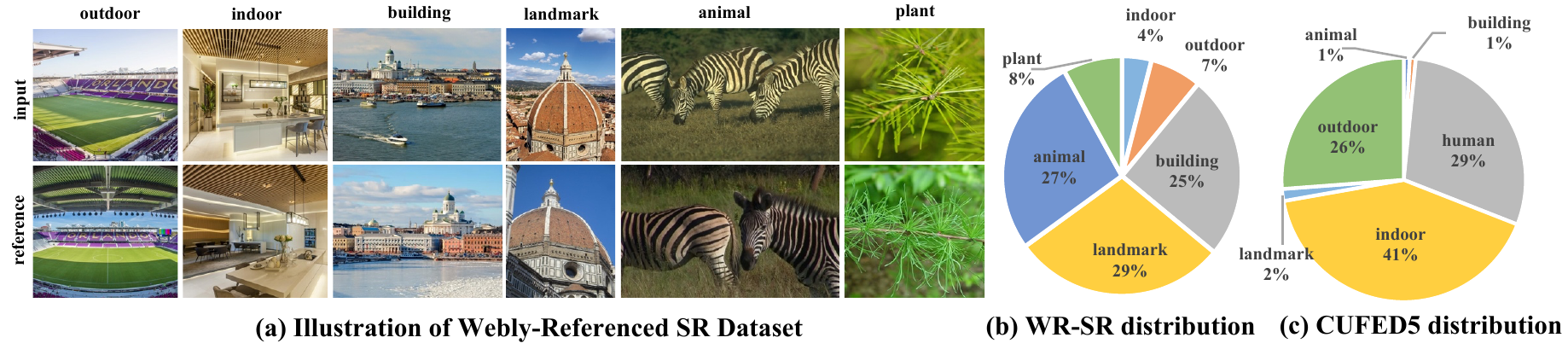}
   \end{center}
  \vspace{-6pt}
   \caption{\textbf{(a) Illustration of Webly-Referenced SR dataset.} The contents of WR-SR dataset include outdoor scenes, indoor scenes, building, famous landmarks, animals and plants. The first line shows the HR input images and the second line is their reference images searched by Google Image. \textbf{(b) WR-SR Dataset Distribution. (c) CUFED5 Dataset Distribution.} The WR-SR dataset contains a more diverse category of images. It has more animal, building and landmark images than CUFED5 dataset. }
   \label{dataset_illustration}
\end{figure*}

\subsection{Dynamic Aggregation Module}

After obtaining correspondences, we fuse textures from reference images by a dynamic aggregation module. 
For every position $p$ in input images, we compute its correspondence point $p^\prime$ in reference images as follows:
\begin{equation}
    p^\prime = \argmax_{q \in I^\prime} \mathrm{cor}_{pq}.
\end{equation}

In order to transfer the texture of reference image, we need to aggregate the information around the position $p^\prime$. We denote $p_{0}$ as the spatial difference between position $p$ and $p^\prime$, \ie $p_{0} = p^\prime - p$. Then the aggregated reference feature $y$ at position $p$ is computed by fusing original reference feature $x$ with a modified DCN as follows:
\begin{equation}
   y(p) = \sum_{k = 1}^{K} w_{k} \cdot x(p + p_{0} + p_{k} + \Delta p_{k}) \cdot \Delta m_{k},
\label{DCN_eq}
\end{equation}
where $p_{k} \in \{(-1, 1), (-1, 0), ..., (1, 1)\}$, $w_{k}$ denotes the convolution kernel weight, $\Delta p_{k}$ and $\Delta m_{k}$ denote the learnable offset and modulation scalar, respectively.

Compared to the reference feature aggregation operation used in \cite{zhang2019image} that cropped patches with a fixed size around corresponding points, our dynamic aggregation module dynamically utilizes the information around the precomputed corresponding points with learnable offset $\Delta p_{k}$.

\subsection{Restoration Module}
The restoration module takes the aggregated reference features and input features as input. These features are first concatenated and then fed into residual blocks to generate the desired output SR image.
We employ the commonly used reconstruction loss $L_{rec}$, perceptual loss $L_{per}$ and adversarial loss $L_{adv}$ for the restoration network. The reconstruction loss we adopted is the $\ell_{1}$-norm. The perceptual loss is calculated on relu5-1 VGG features. 

\begin{table*}
\caption{\textbf{Quantitative Comparisons.} PSNR / SSIM are used for evaluation. We group methods by SISR and Ref-SR. We mark the best results \textbf{in bold}. The models trained with GAN loss are marked in gray. The suffix `-$rec$' means only reconstruction loss is used for training. }
\vspace{-6pt}
\begin{center}
\begin{tabular}{c|l|ccccc}
\Xhline{1pt}
        & Method        & CUFED5        & Sun80   & Urban100  & Manga109     & WR-SR    \\ \hline
\multirow{7}{*}{SISR}   & SRCNN \cite{dong2015image}    & 25.33 / .745    & 28.26 / .781    & 24.41 / .738  & 27.12 / .850   & 27.27 / .767   \\
  & EDSR \cite{lim2017enhanced}   & 25.93 / .777     & 28.52 / .792   & 25.51 / .783   & 28.93 / .891    & 28.07 / .793   \\
  & RCAN \cite{zhang2018image}    & 26.06 / .769     & 29.86 / .810   & 25.42 / .768   & 29.38 / .895    & 28.25 / .799   \\
  & \cellcolor{Gray}SRGAN \cite{ledig2017photo}  & \cellcolor{Gray}24.40 / .702   & \cellcolor{Gray}26.76 / .725    & \cellcolor{Gray}24.07 / .729   & \cellcolor{Gray}25.12 / .802     & \cellcolor{Gray}26.21 / .728   \\
  & ENet \cite{sajjadi2017enhancenet}   & 24.24 / .695  & 26.24 / .702     & 23.63 / .711    & 25.25 / .802   & 25.47 / .699   \\
  & \cellcolor{Gray}ESRGAN \cite{wang2018esrgan}   & \cellcolor{Gray}21.90 / .633    & \cellcolor{Gray}24.18 / .651   & \cellcolor{Gray}20.91 / .620   & \cellcolor{Gray}23.53 / .797    & \cellcolor{Gray}26.07 / .726    \\
  & \cellcolor{Gray}RankSRGAN \cite{zhang2019ranksrgan}    & \cellcolor{Gray}22.31 / .635    & \cellcolor{Gray}25.60 / .667   &  \cellcolor{Gray}21.47 / .624    & \cellcolor{Gray}25.04 / .803     & \cellcolor{Gray}26.15 / .719      \\ \hline \hline
\multirow{11}{*}{Ref-SR} & CrossNet \cite{zheng2018crossnet}    & 25.48 / .764     & 28.52 / .793     & 25.11 / .764    & 23.36 / .741    & -    \\
  & \cellcolor{Gray}SRNTT   & \cellcolor{Gray}25.61 / .764    & \cellcolor{Gray}27.59 / .756    & \cellcolor{Gray}25.09 / .774     & \cellcolor{Gray}27.54 / .862    & \cellcolor{Gray}26.53 / .745       \\
   & SRNTT-$rec$ \cite{zhang2019image}   & 26.24 / .784   & 28.54 / .793   & 25.50 / .783    & 28.95 / .885   & 27.59 / .780     \\
   & \cellcolor{Gray}TTSR   & \cellcolor{Gray}25.53 / .765     & \cellcolor{Gray}28.59 / .774     & \cellcolor{Gray}24.62 / .747    & \cellcolor{Gray}28.70 / .886     & \cellcolor{Gray}26.83 / .762  \\
   & TTSR-$rec$ \cite{yang2020learning}  & 27.09 / .804    & 30.02 / .814     & 25.87 / .784   & 30.09 / .907    & 27.97 / .792   \\
   & \cellcolor{Gray}SSEN    & \cellcolor{Gray}25.35 / .742   & \cellcolor{Gray}-  & \cellcolor{Gray}-   & \cellcolor{Gray}-  & \cellcolor{Gray}-   \\
   & SSEN-$rec$ \cite{Shim_2020_CVPR}  & 26.78 / .791    & -  & -   & -   & -   \\
   & \cellcolor{Gray}E2ENT$^{2}$    & \cellcolor{Gray}24.01 / .705   & \cellcolor{Gray}28.13 / .765   & \cellcolor{Gray}-  & \cellcolor{Gray}-  & \cellcolor{Gray}-  \\
   & E2ENT$^{2}$-$rec$ \cite{xiefeature} & 24.24 / .724   & 28.50 / .789      & -   & -   & -       \\
   & \cellcolor{Gray}CIMR   & \cellcolor{Gray}26.16 / .781    & \cellcolor{Gray}29.67 / .806   & \cellcolor{Gray}25.24 / .778   & \cellcolor{Gray}-    & \cellcolor{Gray}-   \\
   & CIMR-$rec$ \cite{yantowards}   & 26.35 / .789   & 30.07 / .813   & 25.77 / \textbf{.792} & -    & -   \\ \hline \hline
\multirow{2}{*}{Ours}    & \cellcolor{Gray}$C^{2}$-Matching    & \cellcolor{Gray}27.16 / .805   & \cellcolor{Gray}29.75 / .799    & \cellcolor{Gray}25.52 / .764   & \cellcolor{Gray}29.73 / .893   & \cellcolor{Gray}27.80 / .780   \\
   & $C^{2}$-Matching-$rec$   & \textbf{28.24} / \textbf{.841} & \textbf{30.18} / \textbf{.817} & \textbf{26.03} / .785 & \textbf{30.47} / \textbf{.911} & \textbf{28.32} / \textbf{.801} \\
\Xhline{1pt}
\end{tabular}
\end{center}
\vspace{-0.21cm}
\label{quan_comp}
\end{table*}

\subsection{Implementation Details}

The overall network is trained in two-stage: \textbf{1)} training of $C^{2}$-Matching, \ie contrastive correspondence network accompanied with teacher-student correlation distillation. \textbf{2)} training of restoration network.

\noindent\textbf{Training of $C^{2}$-Matching.} We synthesize the image pairs by applying synthetic homography transformation to input images.
Homography transformation matrix is obtained by \textit{cv2.getPerspectiveTransform}.
The margin value $m$ in Eq.~\eqref{loss_margin} is set as $1.0$, the threshold value $T$ in Eq.~\eqref{negative_pair_def} is set as $4.0$, the temperature $\tau$ in Eq.~\eqref{correlation_equa} is set as $0.15$, and the weight $\alpha _{kl}$ for KL-divergence loss is $15$. The learning rate is set as $10^{-3}$. 

\noindent\textbf{Training of Restoration Network.} In this stage, correspondences obtained from the student contrastive correspondence network are used for the calculation of $p_{0}$ specified in Eq.~\eqref{DCN_eq}. The weights for $L_{rec}$, $L_{per}$ and $L_{adv}$ are $1.0$, $10^{-4}$ and $10^{-6}$, respectively. The learning rate for the training of restoration network is set as $10^{-4}$. 
During training, the input sizes for LR images and HR reference images are $40 \times 40$ and $160 \times 160$, respectively.

\section{Webly-Referenced SR Dataset}
\label{section_dataset}

In Ref-SR tasks, the performance relies on similarities between input images and reference images. Thus, the quality of reference images is vital. Currently, Ref-SR methods are trained and evaluated on the CUFED5 dataset \cite{zhang2019image}, where each input image is accompanied by five references of different levels of similarity to the input image. 
A pair of input and reference image in CUFED5 dataset is selected from a same event album. Constructing image pairs from albums ensures a high similarity between the input and reference image. However, in realistic settings, it is not always the case that we can find the reference images from off-the-shelf albums.

In real-world applications, given an LR image, users may find possible reference images through web search engines like Google Image. Motivated by this, we propose a more reasonable dataset named Webly Referenced SR (WR-SR) Dataset to evaluate Ref-SR methods. The WR-SR dataset is much closer to the practical usage scenarios, and it is set up as follows: 

\noindent\textbf{Data Collection.} We select about 150 images from BSD500 dataset \cite{MartinFTM01} and Flickr website. These images are used as query images to search for their visually similar images through Google Image. For each query image, the top 50 similar images are saved as reference image pools for the subsequent Data Cleaning procedure.

\noindent\textbf{Data Cleaning.} Images downloaded from Google Image are of different levels of quality and similarity. Therefore, we manually select the most suitable reference image for each query image. Besides,  since some reference images are significantly larger than input images, we rescale the reference images to a comparable scale as HR input images. We also abandon the images with no proper reference images found.

\noindent\textbf{Data Organization.} A total number of 80 image pairs are collected for WR-SR dataset. Fig. \ref{dataset_illustration} (a) illustrates our WR-SR dataset. The contents of the input images in our dataset include outdoor scenes, indoor scenes, building images, famous landmarks, animals and plants. We analyze the distributions of our WR-SR dataset and CUFED5 dataset. As shown in Fig.~\ref{dataset_illustration} (b), compared to CUFED5 dataset (Fig.~\ref{dataset_illustration} (c)), we have a more diverse category, and we include more animal, landmark, building and plant images.

To summarize, our WR-SR dataset has two advantages over CUFED5 dataset: \textbf{1)} The pairs of input images and reference images are collected in a more realistic way. \textbf{2)} Our contents are more diverse than CUFED5 dataset.

\section{Experiments}

\subsection{Datasets and Evaluation Metrics}

\noindent\textbf{Training Dataset.} We train our models on the training set of CUFED5 dataset \cite{zhang2019image}, which has 11,871 image pairs and each image pair has an input image and a reference image.

\noindent\textbf{Testing Datasets.} The performance are evaluated on the testing set of CUFED5 dataset, SUN80 dataset \cite{sun2012super}, Urban100 dataset \cite{huang2015single}, Manga109 dataset \cite{matsui2017sketch} and our WR-SR dataset. The CUFED5 dataset has 126 input images and each has 5 reference images with different similarity levels. The SUN80 dataset has 80 images with 20 reference images for each input image. WR-SR dataset has been introduced in Section~\ref{section_dataset}. As for the SISR datasets, we adopt the same evaluation setting as \cite{zhang2019image, yang2020learning}. The Urban100 dataset contains 100 building images, and the LR versions of images serve as reference images. The Manga109 dataset has 109 manga images and the reference images are randomly selected from the dataset. 

\noindent\textbf{Evaluation Metrics.} PSNR and SSIM on Y channel of YCrCb space are adopted as evaluation metrics.
Input LR images for evaluation are constructed by bicubic downsampled $4\times$ from HR images.

\begin{figure*}
  \begin{center}
      \includegraphics[width=1.0\linewidth]{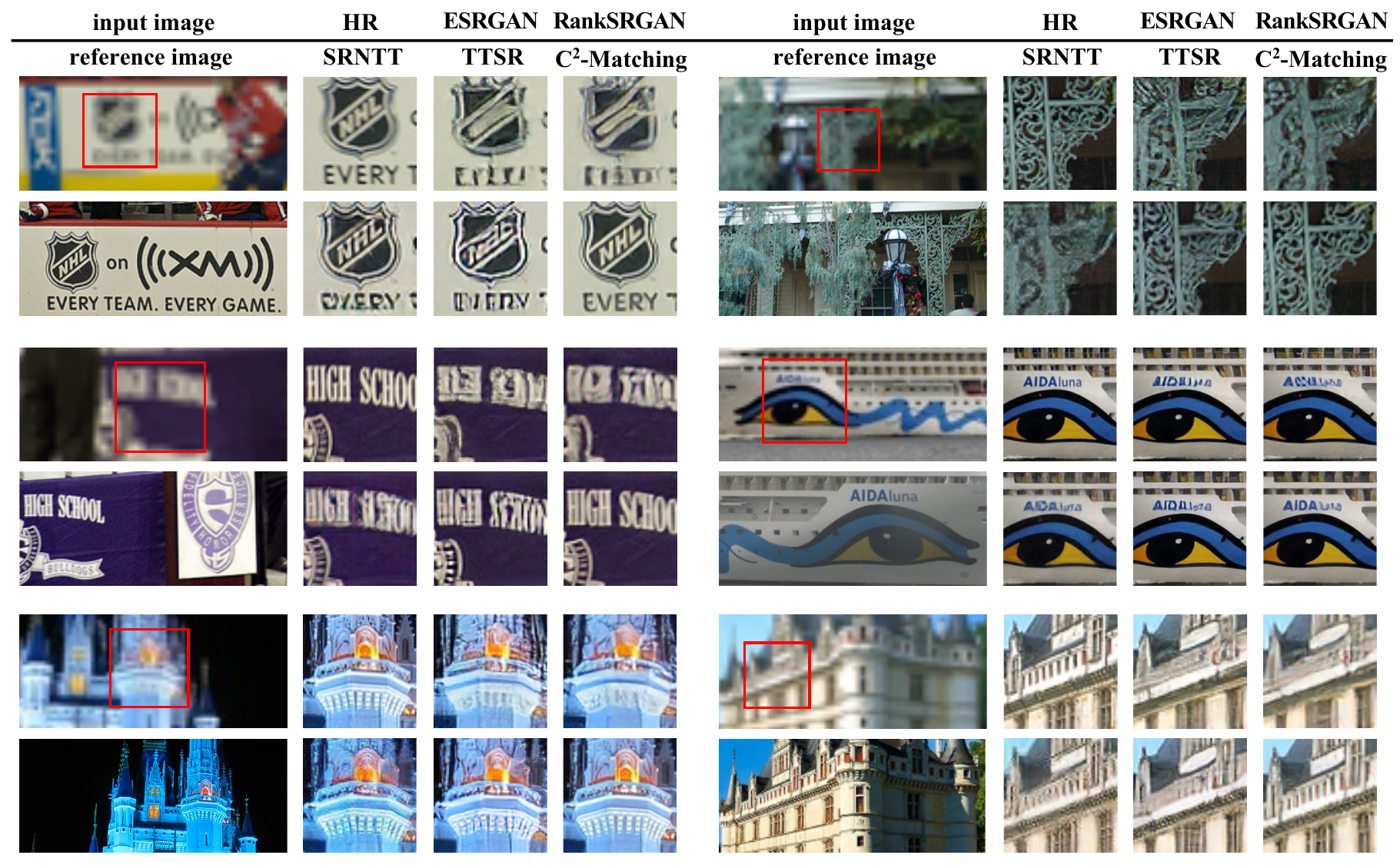}
  \end{center}
  \vspace{-8pt}
  \caption{\textbf{Qualitative Comparisons.} We compare our results with ESRGAN \cite{wang2018esrgan}, RankSRGAN \cite{zhang2019ranksrgan}, SRNTT \cite{zhang2019image}, and TTSR \cite{yang2020learning}. All these methods are trained with GAN loss. Our results have better visual quality with more texture details.}
  \label{visual_comp}
\end{figure*}

\subsection{Results Comparisons}

\noindent\textbf{Quantitative Comparison.} We compare the proposed $C^{2}$-Matching with SISR methods and Ref-SR methods. For SISR methods, we include SRCNN \cite{dong2015image}, EDSR \cite{lim2017enhanced}, RCAN \cite{zhang2018image}, SRGAN \cite{ledig2017photo}, ENet \cite{sajjadi2017enhancenet}, ESRGAN \cite{wang2018esrgan} and RankSRGAN \cite{zhang2019ranksrgan}. For Ref-SR methods, CrossNet \cite{zheng2018crossnet}, SRNTT \cite{zhang2019image}, SSEN \cite{Shim_2020_CVPR}, TTSR \cite{yang2020learning}, E2ENT$^{2}$ \cite{xiefeature} and CIMR \cite{yantowards} are included.

Table~\ref{quan_comp} shows the quantitative comparison results. We mark the methods trained with GAN loss in gray. 
On the standard CUFED5 benchmark, our proposed method outperforms state of the arts by a large margin. We also achieve the best results on the Sun80, Urban100, Manga109 and WR-SR dataset, which demonstrates the great generalizability of $C^{2}$-Matching. Notably, CIMR \cite{yantowards} is a multiple reference-based SR method, which transfers the HR textures from a collection of reference images. Our proposed method performs better than CIMR, which further verifies the superiority of our method. 

\noindent\textbf{Qualitative Evaluation.} Fig.~\ref{visual_comp} shows the qualitative comparison with state of the arts. We compare our method with ESRGAN \cite{wang2018esrgan}, RankSRGAN \cite{zhang2019ranksrgan}, SRNTT \cite{zhang2019image}, and TTSR \cite{yang2020learning}. The results of our method have the best visual quality containing many realistic details and are closer to their respective HR ground-truths. As shown in the top left example, $C^{2}$-Matching successfully recovers the exact word ``EVERY'' while other methods fail. 
Besides, as shown in the bottom left example, our approach can deal with reasonable color and illumination changes to a certain extent without deliberate augmentation.

\begin{figure*}
  \begin{center}
      \includegraphics[width=1.0\linewidth]{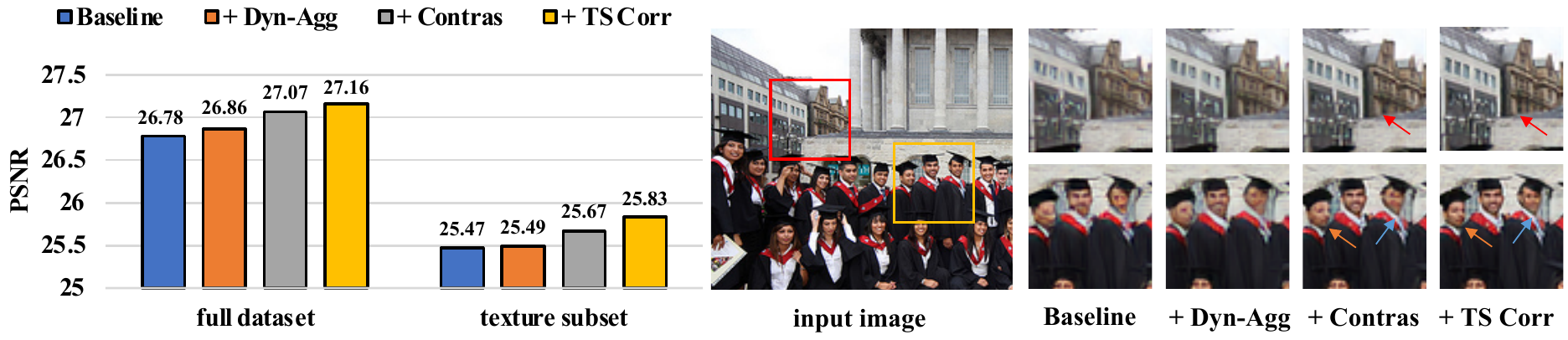}
  \end{center}
  \vspace{-10pt}
  \caption{\textbf{Ablation Study.} We evaluate the effectiveness of each component on the full CUFED5 dataset and the texture region subset. Since the contrastive correspondence (Contras) network and teacher-student correlation distillation (TS Corr) focus on improving texture details, we also add visual comparisons with the component added.}
  \label{ablation_study}
  \vspace{-10pt}
\end{figure*}

\begin{figure*}
  \begin{center}
      \includegraphics[width=1.0\linewidth]{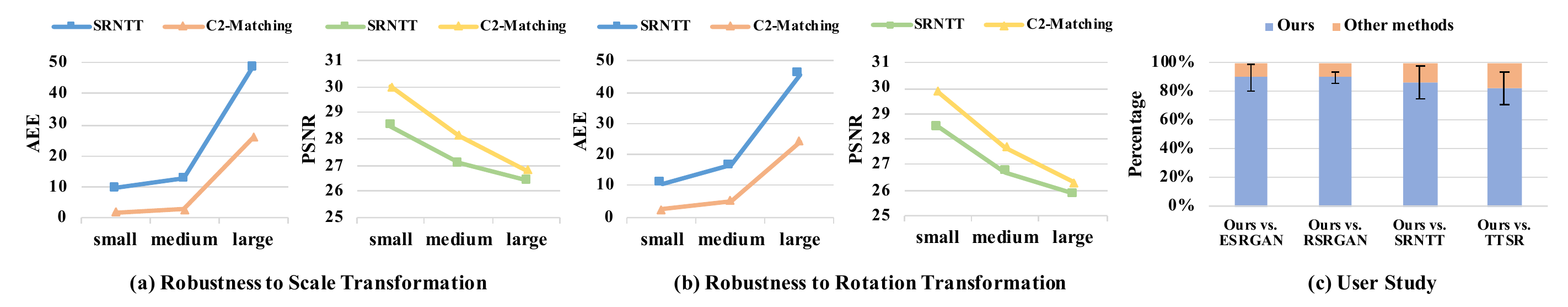}
  \end{center}
  \vspace{-10pt}
  \caption{\textbf{Further Analysis.} (a) Robustness to scale transformation. (b) Robustness to rotation transformations. The proposed $C^{2}$-Matching is more robust to scale and rotation transformation compared to SRNTT. (c) User study. Compared to other state of the arts, over $80\%$ users prefer our results.}
  \label{further_analysis}
  \vspace{-6.5pt}
\end{figure*}

\subsection{Ablation Study}

We perform ablation studies to assess the effectiveness of each module. 
To evaluate the effectiveness of our proposed modules on texture regions, we select a subset of CUFED5 dataset that contains images with complicated textures; we name it ``texture subset".
On top of the baseline model, we progressively add the dynamic aggregation module, contrastive correspondence network and teacher-student correlation distillation to show their effectivenesses. The ablation study results are shown in Fig.~\ref{ablation_study}.

\noindent\textbf{Dynamic Aggregation Module.} We first analyze the effectiveness of the Dynamic Aggregation (Dyn-Agg) module because it deals with the reference texture transfer problem. Only with a better texture transfer module would the improvements of correspondence matching module be reflected, \ie the baseline accompanied with Dyn-Agg module provides a stronger backup. The Dyn-Agg module dynamically fuses the HR information from reference images. Compared to the previous scheme that cropped patches of a fixed size from HR reference features, the Dyn-Agg module has a more flexible patch size. With Dyn-Agg module, we observe an increment in PSNR by 0.08dB in the full dataset. 

\noindent\textbf{Contrastive Correspondence Network.} We further replace the fixed VGG feature matching module with the contrastive correspondence (Contras) network. With the learnable contrastive correspondence network, the PSNR value increases by about 0.2dB.
This result demonstrates the contrastive correspondence network computes more accurate correspondences and further boosts the performance of restoration. Fig.~\ref{ablation_study} shows one example of visual comparisons. With the contrastive correspondence network, the output SR images have more realistic textures.

\noindent\textbf{Teacher-Student Correlation Distillation.} With the teacher-student correlation (TS Corr) distillation, the performance further increases by 0.09dB on the whole dataset. For the texture subset, the performance increases by 0.16dB. The TS Corr module aims to push closer the correspondence of LR-HR student Contras network and that of HR-HR teacher Contras network. Since HR-HR teacher matching model is more capable of matching texture regions, the TS Corr module mainly boosts the performance on texture regions. As indicated in Fig.~\ref{ablation_study}, with TS Corr module, the textures of output SR images are enriched.

\subsection{Further Analysis}

\noindent\textbf{Robustness to Scale and Rotation Transformations.} We perform further analysis on the robustness of our $C^{2}$-Matching to scale and rotation transformations. We build a transformation-controlled dataset based on CUFED5 dataset. The scaled and rotated versions of input images serve as reference images. 
We adopt two metrics to measure the robustness: Average End-to-point Error (AEE) for matching accuracy and PSNR for restoration performance.

Fig.~\ref{further_analysis} shows the robustness to the scale and transformations in AEE and PSNR. We separately analyze the impact of scale and rotation. We classify the degrees of scale and rotations into three groups: small, medium and large. The AEE rises as the degrees of transformations increases, which indicates a larger degree of transformation makes it harder to perform correspondence matching. Based on AEE, our proposed $C^{2}$-Matching computes more accurate correspondences than SRNTT under scale and rotation transformations. With the features that are more robust to scale and rotation transformations, according to the PSNR, the restoration performance of our proposed $C^{2}$-Matching is also more robust than that of SRNTT \cite{zhang2019image}. It should be noted that large transformations are not included during training but our proposed $C^{2}$-Matching still exhibits superior performance compared to SRNTT.

\noindent\textbf{User Study.} We perform a user study to further demonstrate the superiority of our method qualitatively. A total number of 20 users are asked to compare the visual quality of our method and state of the arts on the CUFED5 dataset, including ESRGAN \cite{wang2018esrgan}, RankSRGAN \cite{zhang2019ranksrgan}, SRNTT \cite{zhang2019image} and TTSR \cite{yang2020learning}. We present images in pairs, of which one is the result of our method, and ask users to choose the one offering better visual quality. As shown in Fig.~\ref{further_analysis}, over 80\% of the users felt that the result of our method is superior compared to that of state of the arts.

\section{Conclusion}

In this paper, we propose a novel $C^{2}$-Matching for robust reference-based super-resolution, which consists of contrastive correspondence network, teacher-student correlation distillation and dynamic aggregation module. 
The motivation of contrastive correspondence network is to perform scale and rotation robust matching between input images and reference images. 
The teacher-student correlation distillation is proposed to distill the teacher HR-HR matching to guide the student LR-HR matching to improve the visual quality of texture regions. 
After obtaining the correspondences, we fuse the information of reference images through a dynamic aggregation module. 
We achieve over 1dB improvement in PSNR over state of the arts. To facilitate a more realistic evaluation of Ref-SR tasks, we also contribute a new benchmark named WR-SR dataset, which is collected in a more realistic way.

\noindent\textbf{Acknowledgement}.
This research was conducted in collaboration with SenseTime. This work is supported by NTU NAP and A*STAR through the Industry Alignment Fund - Industry Collaboration Projects Grant.

{\small
\bibliographystyle{ieee_fullname}
\bibliography{egbib}
}
\newpage
\appendix
\section*{Supplementary}
In this supplementary file, we will explain the network structures (\ie Contrastive Correspondence Network and Restoration Network) and training details in Section~\ref{structure}. Then we will introduce training losses we used in Section~\ref{loss_function}. In Section~\ref{param_number}, the model size will be analyzed. In Section~\ref{vis_comp}, we will provide more visual comparisons with state-of-the-art methods. Finally, we will show more visual comparisons of ablation study in Section~\ref{section_ablation}.

\section{Network Structures and Training Details}
\label{structure}
\subsection{Contrastive Correspondence Network}

\noindent\textbf{Network Structure.} Table~\ref{table_feature_extractor} shows the detailed feature extractor structure of contrastive correspondence network. Since the resolutions of input image and reference image are different, we adopt two feature extractors for LR input image and HR reference image, respectively. 

\vspace{6pt}

\makeatletter\def\@captype{table}\makeatother
\caption{\textbf{The feature extractor structure of contrastive correspondence network.} The kernel size of convolution layers is $3 \times 3$ and the MaxPool operation is with kernel size of $2 \times 2$.}
\begin{center}
\begin{tabular}{l|l}
\Xhline{1pt}
\textbf{\#}  & \textbf{Layer name(s)} \\ \hline \hline
 0 & Conv (3, 64), ReLU      \\ \hline
 1 & Conv (64, 64), ReLU  \\ \hline
 2 & MaxPool (2 $\times$ 2) \\ \hline
 3 & Conv (64, 128), ReLU  \\ \hline
 4 & Conv (128, 128), ReLU  \\ \hline
 5 & MaxPool (2 $\times$ 2) \\ \hline
 6 & Conv (128, 256)   \\ \hline
\Xhline{1pt}
\end{tabular}
\label{table_feature_extractor}
\end{center}

\noindent\textbf{Training Details.} 
To enable teacher-student correlation distillation, a teacher contrastive correspondence network should be first trained. The hyperparameters for the training of teacher model are set as follows: the margin value $m$ is $1.0$, the threshold value $T$ is $4.0$, the batch size is set as $8$, and the learning rate is $10^{-3}$. We use the pretrained weights of VGG-16 to initialize the feature extractor.  Then the student contrastive correspondence network is trained with the teacher network fixed. The margin value $m$, threshold value $T$, batch size and learning rate are the same as the teacher network. The temperature value $\tau$ is $0.15$, and the weight $\alpha_{kl}$ for KL-divergence loss is $15$. 

\subsection{Restoration Network}

\noindent\textbf{Network Structure.} The restoration network consists of dynamic aggregation module and restoration module. For each image, three reference features (\ie pretrained VGG relu3\_1, relu2\_1, relu1\_1 feature \cite{simonyan2014very}) are aggregated by dynamic aggregation module, and the aggregated reference features are denoted as Aggregated Reference Feature1, Aggregated Reference Feature2 and Aggregated Reference Feature3, respectively. 
The structure of restoration module is illustrated in Table.~\ref{table_restoration_network}.

\vspace{6pt}
\makeatletter\def\@captype{table}\makeatother
\caption{\textbf{The structure of restoration module.} The kernel size of convolution layers is $3 \times 3$. PixelShuffle layers are 2$\times$. RB denotes residual block. Aggregated Reference Feature denotes the reference feature aggregated by the dynamic aggregation module.}
\begin{center}
\begin{tabular}{l|l}
\Xhline{1pt}
\textbf{\#}  & \textbf{Layer name(s)} \\ \hline \hline
 0 & Conv(3, 64), LeakyReLU \\ \hline
 1 & RB [Conv(64, 64), ReLU, Conv(64, 64)] $\times$ 16 \\ \hline
 2 & Concat [\#1, Aggregated Reference Feature1]  \\ \hline
 3 & Conv(320, 64), LeakyReLU  \\ \hline
 4 & RB [Conv(64, 64), ReLU, Conv(64, 64)] $\times$ 16  \\ \hline
 5 & ElementwiseAdd(\#1, \#4) \\ \hline
 6 & Conv(64, 256), PixelShuffle, LeakyReLU  \\ \hline
 7 & Concat [\#6, Aggregated Reference Feature2]  \\ \hline
 8 & Conv(192, 64), LeakyReLU  \\ \hline
 9 & RB [Conv(64, 64), ReLU, Conv(64, 64)] $\times$ 16 \\ \hline
 10 & ElementwiseAdd(\#6, \#9)  \\ \hline
 11 & Conv(64, 256), PixelShuffle, LeakyReLU \\ \hline
 12 & Concat [\#11, Aggregated Reference Feature3]  \\ \hline
 13 & Conv(128, 64), LeakyReLU \\ \hline
 14 & RB [Conv(64, 64), ReLU, Conv(64, 64)] $\times$ 16  \\ \hline
 15 & ElementwiseAdd(\#11, \#14)  \\ \hline
 16 & Conv(64, 32), LeakyReLU  \\ \hline
 17 & Conv(32, 3) \\ \hline
 
\Xhline{1pt}
\end{tabular}
\label{table_restoration_network}
\end{center}

\noindent\textbf{Training Details.} 
The learning rate is set as $10^{-4}$. For the training of the network with adversarial loss and perceptual loss, we adopt the same setting as \cite{zhang2019image} (\ie the network is trained with only reconstruction loss for the first 10K iterations).

\section{Loss Functions}
\label{loss_function}

\noindent\textbf{Reconstruction Loss.} The $\ell_1$-norm is adopted to keep the spatial structure of the LR images. It is defined as follows:
\begin{equation}
   L_{rec} = \left \| I^{HR} - I^{SR} \right \|_{1}.
\end{equation}

\noindent\textbf{Perceptual Loss.} The perceptual loss \cite{johnson2016perceptual} is employed to improve the visual quality. It is defined as follows:
\begin{equation}
    L_{per} = \frac{1}{V}\sum_{i=1}^{C}\left \| \phi _{i} (I^{HR}) - \phi _{i} (I^{SR})  \right \|_{F},
\end{equation}
where $V$ and $C$ denotes the volume and channel number of feature maps. $\phi$ denotes the relu5\_1 features of VGG19 model \cite{simonyan2014very}. $\left \| \cdot  \right \|_{F}$ denotes the Frobenius norm.

\noindent\textbf{Adversarial Loss.} The adversarial loss \cite{ledig2017photo} is defined as follows:
\begin{equation}
    L_{adv} = - D(I^{SR}).
\end{equation}
The loss for training discriminator $D$ is defined as follows:
\begin{equation}
    L_{D} = D(I^{SR}) - D(I^{HR}) + \lambda (\left \| \nabla _{\hat{I}} D(\hat{I}) \right \|_{2} - 1)^2.
\end{equation}
where $\hat{I}$ is the random convex combination of $I^{SR}$ and $I^{HR}$.

\section{Comparison of Model Size}
\label{param_number}
The comparison of model size (\ie the number of trainable parameters) is shown in Table~\ref{table_param_comp}. Our proposed $C^{2}$-Matching has a total number of 8.9M parameters and achieves a PSNR of 28.24dB. For a fair comparison in terms of model size, we build a light version of $C^{2}$-Matching, which has fewer trainable parameters. The $C^{2}$-Matching-$light$ is built by setting the number of residual blocks of layer \#9 and layer \#14 to 8 and 4, respectively, and removing the Aggregated Reference Feature1. The $C^{2}$-Matching-$light$ has a total number of 4.8M parameters. The light version has fewer parameters than TTSR \cite{yang2020learning} but significantly better performance.

\vspace{6pt}
\makeatletter\def\@captype{table}\makeatother
\caption{\textbf{Model sizes of different methods.} PSNR / SSIM are adopted as the evaluation metrics.}
\begin{center}
\begin{tabular}{l|l|l}
\Xhline{1pt}
\textbf{Method}  & \textbf{Params} & \textbf{PSNR/SSIM} \\ \hline \hline
 RCAN \cite{zhang2018image} & 16M & 26.06 / .769 \\ \hline
 RankSRGAN \cite{zhang2019ranksrgan} & 1.5M     & 22.31 / .635 \\ \hline
 CrossNet \cite{zheng2018crossnet} & 33.6M & 25.48 / .764 \\ \hline
 SRNTT \cite{zhang2019image} & 4.2M & 26.24 / .784 \\ \hline
 TTSR \cite{yang2020learning} & 6.4M & 27.09 / .804 \\ \hline
 $C^{2}$-Matching-$light$ & 4.8M  & 28.14 / .839 \\ \hline
 $C^{2}$-Matching & 8.9M & 28.24 / .841 \\ \hline
\Xhline{1pt}
\end{tabular}
\label{table_param_comp}
\end{center}



\section{More Visual Comparisons with State-of-the-art Methods}
\label{vis_comp}

\begin{figure*}
  \begin{center}
      \includegraphics[width=1.0\linewidth]{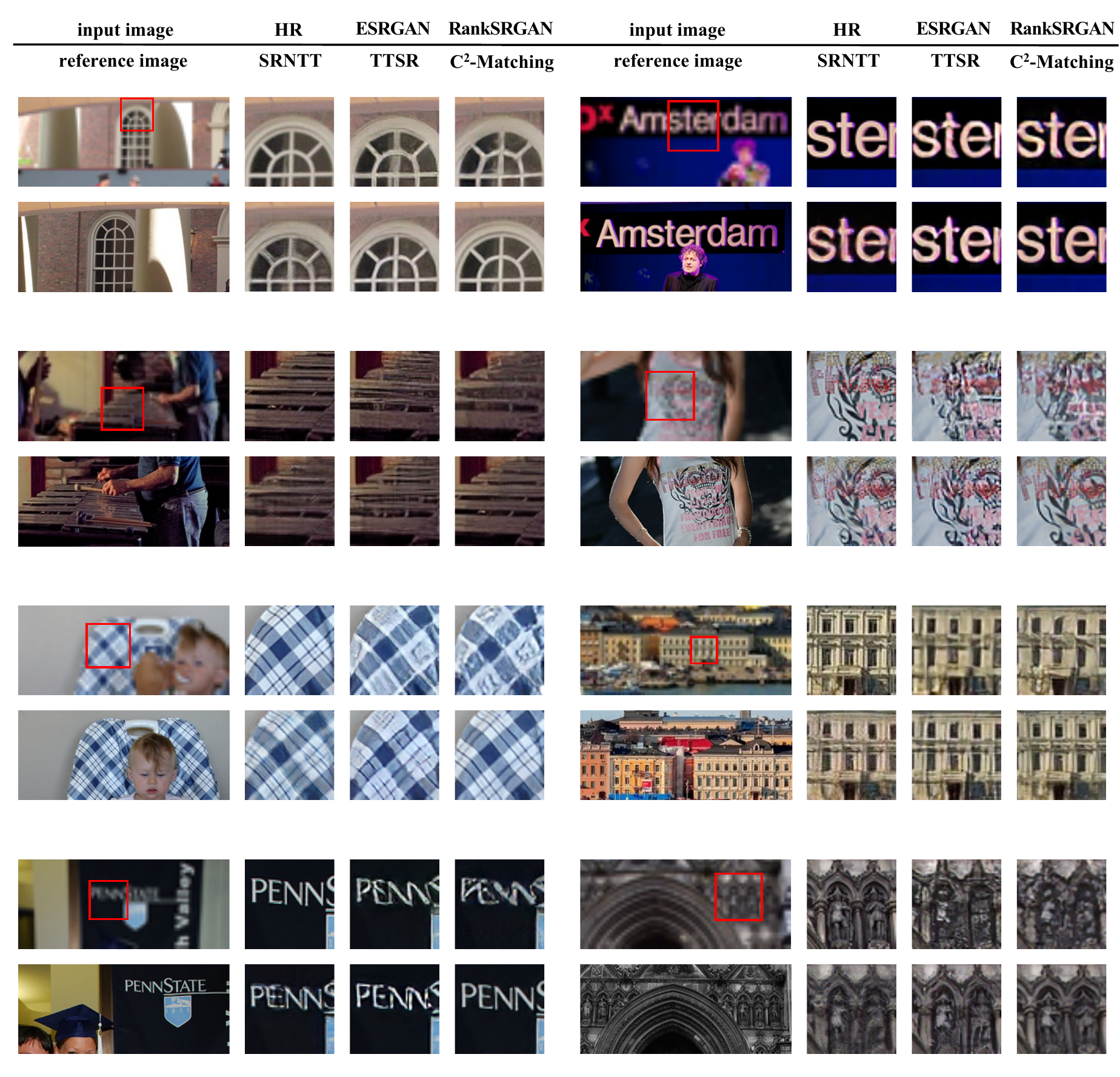}
  \end{center}
  \caption{\textbf{More visual comparisons.} We compare our results with ESRGAN \cite{wang2018esrgan}, RankSRGAN \cite{zhang2019ranksrgan}, SRNTT \cite{zhang2019image}, and TTSR \cite{yang2020learning}. All these methods are trained with GAN loss. Our results have better visual quality with more texture details.}
  \label{visual_comp1}
\end{figure*}

\begin{figure*}
  \begin{center}
      \includegraphics[width=1.0\linewidth]{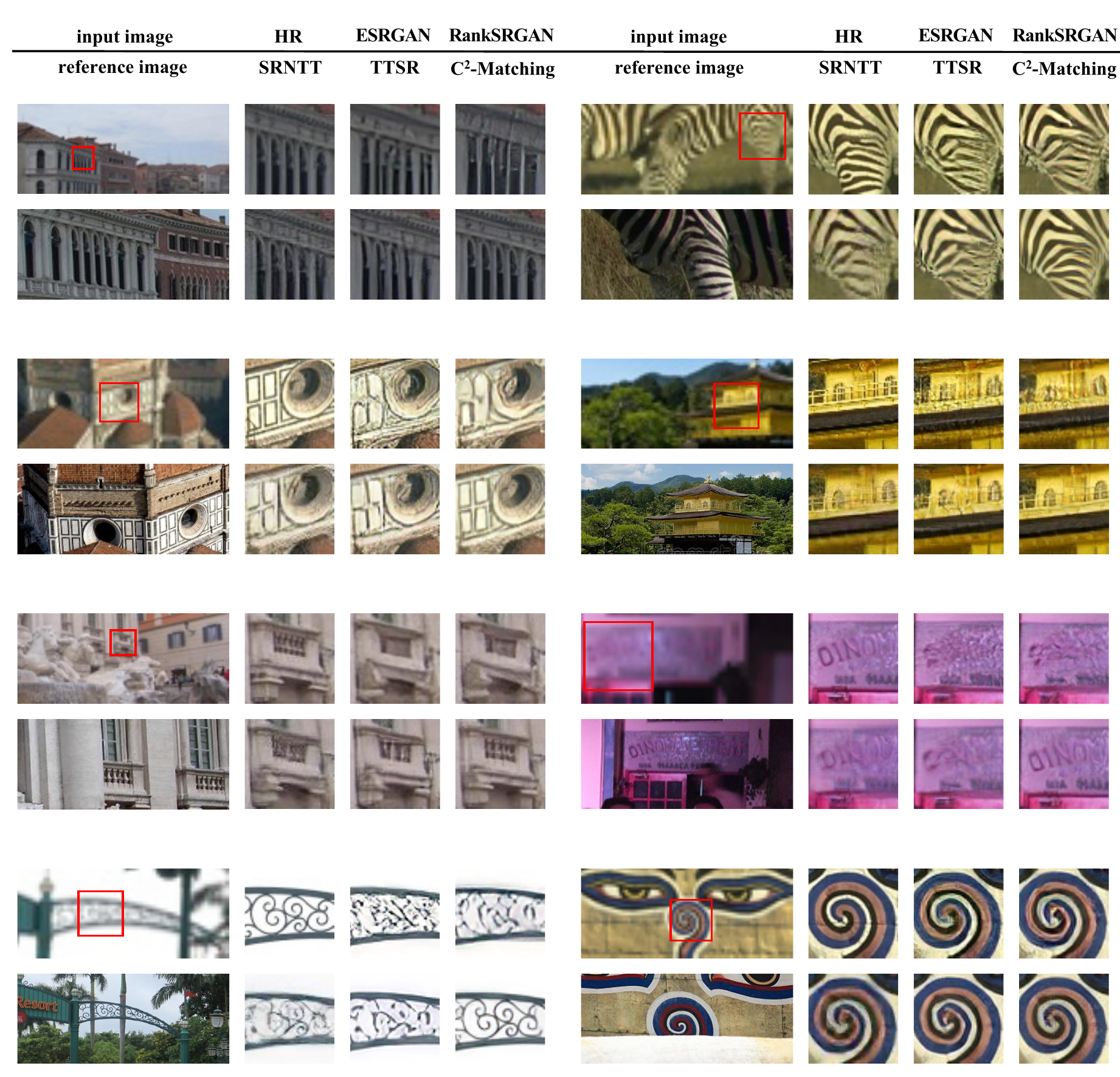}
  \end{center}
  \caption{\textbf{More visual comparisons.} We compare our results with ESRGAN \cite{wang2018esrgan}, RankSRGAN \cite{zhang2019ranksrgan}, SRNTT \cite{zhang2019image}, and TTSR \cite{yang2020learning}. All these methods are trained with GAN loss. Our results have better visual quality with more texture details.}
  \label{visual_comp2}
\end{figure*}

In Fig.~\ref{visual_comp1} and Fig.~\ref{visual_comp2}, more visual comparisons with ESRGAN \cite{wang2018esrgan}, RankSRGAN \cite{zhang2019ranksrgan}, SRNTT \cite{zhang2019image} and TTSR \cite{yang2020learning} are provided. The images restored by our proposed $C^{2}$-Matching have better visual quality.

\section{More Visual Comparisons of Ablation Study}
\label{section_ablation}

\begin{figure*}
  \begin{center}
      \includegraphics[width=1.0\linewidth]{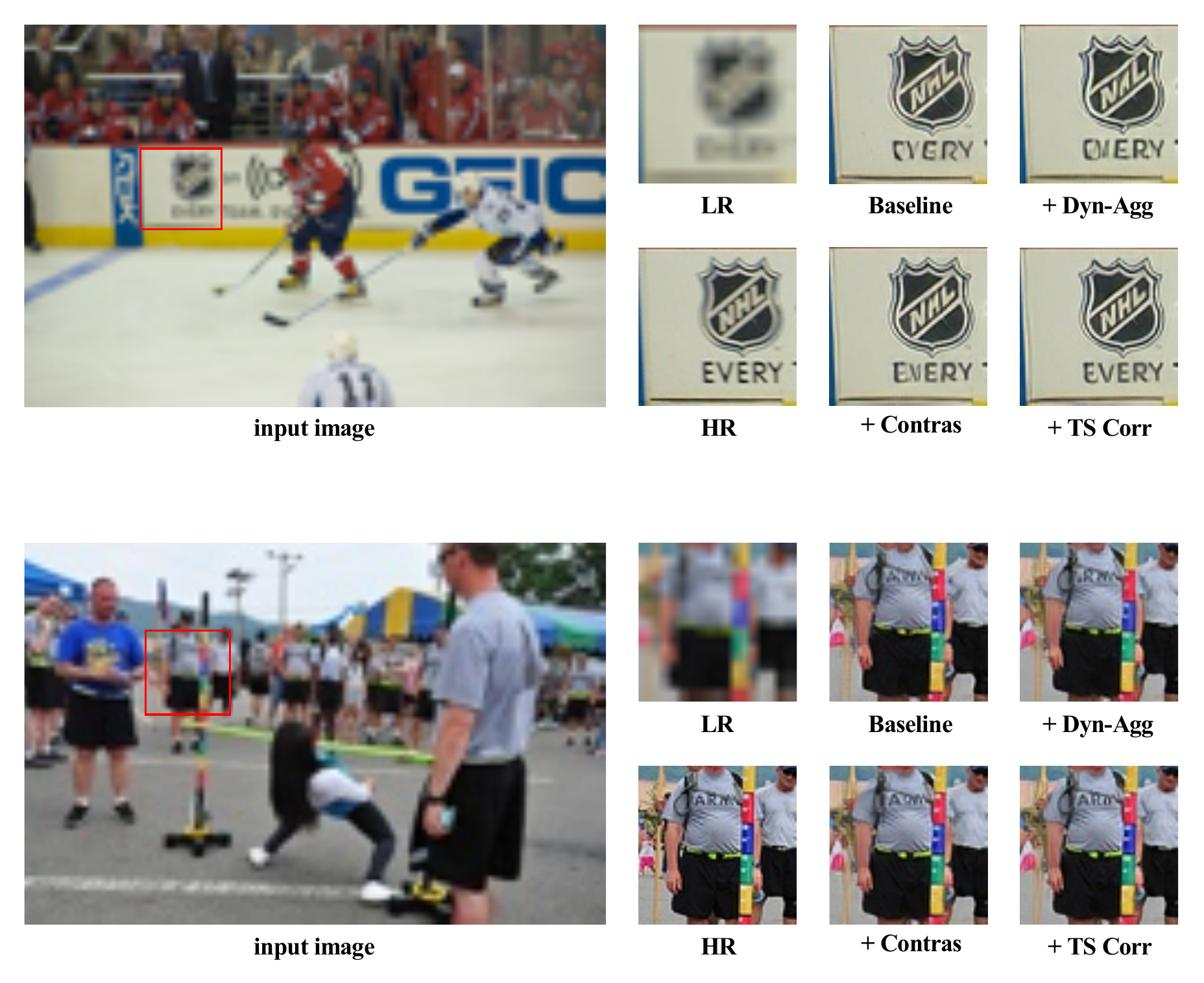}
  \end{center}
  \caption{\textbf{More visual comparisons of ablation study.} On top of the baseline model, Dynamic Aggregation Module (Dyn-Agg), Contrastive Correspondence Network (Contras) and Teacher-Student Correlation Distillation (TS Corr) are progressively added.}
  \label{more_ablation}
\end{figure*}

In this paper, the proposed $C^{2}$-Matching consists of three major components: Dynamic Aggregation Module (Dyn-Agg), Contrastive Correspondence Network (Contras) and Teacher-Student Correlation Distillation (TS Corr). On top of the baseline model, we progressively add the Dyn-Agg module,  Contras network and TS Corr distillation. In Fig.~\ref{more_ablation}, we show more visual comparisons with these proposed modules progressively added.

\end{document}